\documentclass[final,5p,times,twocolumn,authoryear]{elsarticle}

\usepackage{amssymb}
\usepackage{amsmath}
\usepackage{hyperref}
\usepackage{color}

\usepackage{tabularx}
\usepackage[table,xcdraw]{xcolor}
\usepackage{CJKutf8}
\usepackage{tabularx}
\usepackage{fontawesome}
\usepackage{booktabs}
\usepackage{multirow}

\journal{Elsevier}

\begin{document}
\begin{CJK}{UTF8}{gbsn}

\begin{frontmatter}



\title{Can large language models understand uncommon meanings of common words?} 

\author[label1]{Jinyang Wu}
\ead{wu-jy23@mails.tsinghua.edu.cn}
\author[label1]{Feihu Che}
\ead{qkr@mail.tsinghua.edu.cn}
\author[label2]{Xinxin Zheng}
\ead{zhengxinxin2021@ia.ac.cn}
\author[label1]{Shuai Zhang}
\ead{zhang_shuai@mail.tsinghua.edu.cn}
\author[label1]{Ruihan Jin}
\ead{jrh20@mails.tsinghua.edu.cn}
\author[label2]{Shuai Nie}
\ead{nss90221@gmail.com}
\author[label1]{Pengpeng Shao}
\ead{ppshao@mail.tsinghua.edu.cn}
\author[label1,label3]{Jianhua Tao\corref{cor1}}
\ead{jhtao@tsinghua.edu.cn}

\cortext[cor1]{Corresponding author}

\affiliation[label1]{organization={Department of Automation, Tsinghua University},
            city={Beijing},
            postcode={100084}, 
            country={China}}
\affiliation[label2]{organization={Institute of automation, Chinese academy of science},
            city={Beijing},
            postcode={100190}, 
            country={China}}
\affiliation[label3]{organization={Beijing National Research Center for Information Science and Technology, Tsinghua University},
            city={Beijing},
            postcode={100084}, 
            country={China}}


\begin{abstract}
Large language models (LLMs) like ChatGPT have shown significant advancements across diverse natural language understanding (NLU) tasks, including intelligent dialogue and autonomous agents. Yet, lacking widely acknowledged testing mechanisms, answering `whether LLMs are stochastic parrots or genuinely comprehend the world' remains unclear, fostering numerous studies and sparking heated debates. Prevailing research mainly focuses on surface-level NLU, neglecting fine-grained explorations. However, such explorations are crucial for understanding their unique comprehension mechanisms, aligning with human cognition, and finally enhancing LLMs' general NLU capacities. To address this gap, our study delves into LLMs' nuanced semantic comprehension capabilities, particularly regarding common words with uncommon meanings. The idea stems from foundational principles of human communication within psychology, which underscore accurate shared understandings of word semantics. Specifically, this paper presents the innovative construction of a \textbf{Le}xical \textbf{S}emantic \textbf{C}omprehension (LeSC) dataset with novel evaluation metrics, the first benchmark encompassing both fine-grained and cross-lingual dimensions. Introducing models of both open-source and closed-source, varied scales and architectures, our extensive empirical experiments demonstrate the inferior performance of existing models in this basic lexical-meaning understanding task. Notably, even the state-of-the-art LLMs GPT-4 and GPT-3.5 lag behind 16-year-old humans by 3.9$\%$ and 22.3$\%$, respectively. Additionally, multiple advanced prompting techniques and retrieval-augmented generation are also introduced to help alleviate this trouble, yet limitations persist. By highlighting the above critical shortcomings, this research motivates further investigation and offers novel insights for developing more intelligent LLMs. The resources are available at \href{https://github.com/jinyangwu/LeSC}{https://github.com/jinyangwu/LeSC.}

\end{abstract}




\end{frontmatter}

\section{Introduction}
\label{sec1}
\noindent``\textit{Any fool can know. The point is to understand.}''\newline 
\rightline{--- \textit{Albert Einstein}}

Researchers in AI community, especially Natural Language Processing (NLP), have been investigating the fundamental principles of intelligence for years\textcolor[RGB]{0,128,172}{~\citep{bubeck2023sparks}}. Benefitting from unprecedented scales of model size and training corpus, LLMs like ChatGPT\textcolor[RGB]{0,128,172}{~\citep{open2023introducing}}, Gemini\textcolor[RGB]{0,128,172}{~\citep{team2023gemini}}, and LLaMA\textcolor[RGB]{0,128,172}{~\citep{touvron2023llama}} have shown surprising, even emergent Natural Language Understanding (NLU) capabilities\textcolor[RGB]{0,128,172}{~\citep{Daking-AAAI23}} at or surpassing human levels even in unseen scenarios, which ensures proficient execution across diverse downstream tasks, such as sentiment analysis\textcolor[RGB]{0,128,172}{~\citep{wang-etal-2018-glue}}, question answering\textcolor[RGB]{0,128,172}{~\citep{ijcai2023p457}}, and autonomous agents\textcolor[RGB]{0,128,172}{~\citep{chen2024agentverse}}.

As Albert Einstein once stated, ``Any fool can know. The point is to understand''\footnote{Words pronounced by Albert Einstein in the TV play “Doctor Einstein Before Lunch” by Ernest Kinoy, first aired on US
NBC television on 20 May 1973. https://falschzitate.blogspot.com/2021/10/jeder-dummkopf-kann-es-wissen-der-punkt.html.}. From a human cognitive perspective, understanding is crucial for possessing human intelligence rather than relying solely on memorization and mimicry\textcolor[RGB]{0,128,172}{~\citep{lonergan1957insight,sternberg1983components}}. Given LLMs' impressive performance across diverse semantic understanding tasks, there is no doubt that LLMs are powerful tools and possess some degree of language understanding and intelligence\textcolor[RGB]{0,128,172}{~\citep{haggstrom2023large}}. Nevertheless, recent studies\textcolor[RGB]{0,128,172}{~\citep{bender2021dangers,bubeck2023sparks,xu2024academically}} have expressed considerable concerns about the scientific and also philosophical question: ``whether LLMs genuinely understand the world or just mimic language patterns and logic, i.e. stochastic parrots''. For example, \textcolor[RGB]{0,128,172}{\citep{bender2021dangers}} advocates that NLP researchers carefully examine the risks of blindly pursuing larger models, considering factors such as resource consumption and the significance of model outputs due to their inherent limitations. \textcolor[RGB]{0,128,172}{~\citep{borji2023stochastic}} points out that LLMs may excel in understanding abstract tasks detached from the physical world (e.g. mathematics or coding) but struggle with more real-world understandings, which is essential for developing more intelligent LLMs. \textcolor[RGB]{0,128,172}{~\citep{li2023dark}} discusses that new legal and ethical risks are emerging due to stochastic parrots and hallucinations.

Numerous attempts have been made to answer the above question, especially in creating meaningful benchmarks that reflect LLMs' ability to accurately grasp human-conveyed semantics. \textcolor[RGB]{0,128,172}{\citep{choi-etal-2023-llms}} introduced a theory-driven benchmark and measured how well LLMs understand social language at the sentence level; \textcolor[RGB]{0,128,172}{\citep{riccardi2023two}} proposed a noun-noun combination-based dataset to assess LLMs' capacity for meaningfulness judgment, which requires a high level of language understanding; \textcolor[RGB]{0,128,172}{\citep{jang2023can}} explored whether LLMs actually understand what they are instructed to do facing word-level modifications; GLUE\textcolor[RGB]{0,128,172}{~\citep{wang-etal-2018-glue}} aimed to evaluate the performance of NLP models on eight language understanding tasks, such as Duplicate Sentence Detection; \textcolor[RGB]{0,128,172}{~\citep{zheng-etal-2022-fewnlu}} constructed FewNLU to makes a systematic assessment of the few-shot learning problems, and Promptbench\textcolor[RGB]{0,128,172}{~\citep{zhu2023promptbench}} were proposed to analyze LLMs' robustness while facing NLU tasks with attacks. However, they primarily focus on sentence-level semantic understanding or word-level instruction comprehension, overlooking nuanced dimensions that hinder our holistic comprehension of their linguistic competence. Moreover, the basis of communication in psychology relies on a shared understanding of word meanings\textcolor[RGB]{0,128,172}{~\citep{miller1967psychology,de2023common}}, a fine-grained language comprehension capacity. Therefore, proposing a targeted dataset to bridge this gap is of profound significance.

In this work, we aim to comprehensively discuss the above challenge question focusing on uncommon meanings of common words, a fine-grained perspective in line with communication psychology and human cognition\textcolor[RGB]{0,128,172}{~\citep{clark1983understanding,duvivier1999common}}. Motivated by this intuitive inspiration, we construct the LeSC dataset to assess the word-level NLU ability, and introduce two metrics including our proposed weighted accuracy, which effectively mitigates the impact of model preferences, thus fostering a more fair and objective evaluation. Contrary to prior work, our proposed dataset LeSC specializes in context-aware fine-grained lexical semantics understanding (LSU, definition in theoretical linguistics\textcolor[RGB]{0,128,172}{~\citep{geeraerts2002theoretical}}), and also provides cross-lingual transfer tests, which delve deeper into the model's comprehension at a more precise semantic level. Given this benchmark, a series of confirmatory experimental research can be conducted, and the results highlight a prevailing lack of LSU capability among current LLMs, with even ChatGPT producing less-than-satisfactory performance (\textcolor[RGB]{0,128,172}{Figure \ref{figure1}}). We then take a step towards investigating the language capability transfer in LLMs. Moreover, some strategies are also discussed to effectively alleviate the above problem, such as few-shot prompting\textcolor[RGB]{0,128,172}{~\citep{brown2020language}} and chain-of-thought\textcolor[RGB]{0,128,172}{~\citep{wei2022chain}}. For in-depth analysis, we also posit conjectures to dive into the underlying explanations and magic through additional studies. To sum up, the key contributions are summarized as follows:

\begin{figure}[ht]
\centering
\includegraphics[width=1.0\linewidth]{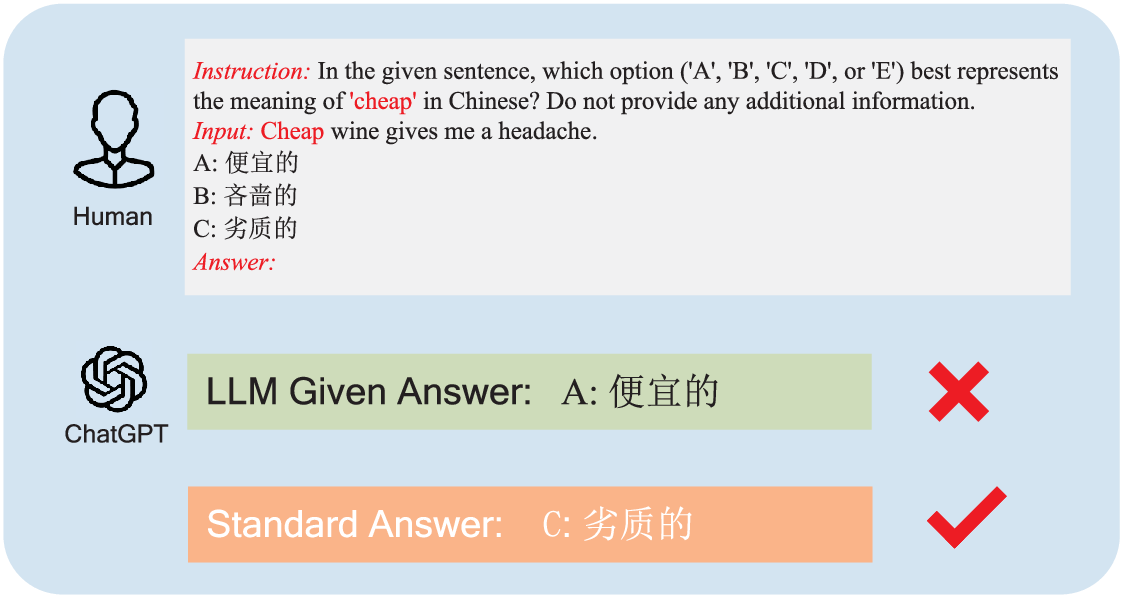}
\caption{An example from LeSC dataset. Within the gray box are the inputs, comprising a prompt, a question, and provided options, and 'A', 'B', 'C' refer to 'low in price', 'unwilling to spend money', 'of poor quality; inferior', respectively. Within the green box, the answer of ChatGPT is 'A', inconsistent with the correct answer 'C'.}\label{figure1}
\end{figure}

\begin{itemize}
    \item We introduce LeSC, a pioneering open-source benchmark with a fair and objective evaluation metric rooted in model preferences for assessing the fine-grained LSU capability of LLMs. The design of LeSC ensures its validity, thereby furnishing a valuable foundation for future alignment with human comprehension and facilitating other NLU-based research like robotic navigation with LLMs. 

    \item Extensive experiments reveal that existing models, affected by their inherent imitation tendencies, overconfidence, and inadequate language capacity transfer, exhibit limited proficiency in the basic LSU task, despite their notable efficacy in more complex NLU tasks.

    \item Comparing ChatGPT (GPT-3.5 and GPT-4) with the human performance measured by asking 16-year-old humans from diverse backgrounds on the same task, we show that there is a huge ($\sim$22.3$\%$ and $\sim$3.9$\%$) performance gap to close. 

    \item Results indicate that advancing prompting techniques and retrieval-augmented generation partially help mitigate this challenge, however, their benefits tend to diminish or even become counterproductive on very large language models.

    \item Some visualizations are presented to help delve into the underlying explanations and mechanisms.
    
\end{itemize}

Exploring the fine-grained LSU ability of LLMs can offer crucial insights into NLP research, consequently leading to a comprehensive enhancement in the efficacy of various downstream tasks like translation and summarization. Our findings indicate that despite remarkable performance on challenging NLU tasks, state-of-the-art (SOTA) language models like ChatGPT still have fundamental weaknesses in lexical context understanding and extracting proper meanings from the input. Beyond solving more challenging problems, future work should also take into account the underestimated challenge of fine-grained LSU, thus offering guidance and assistance in developing more powerful supermodels.

\begin{figure*}[ht]
\centering
\includegraphics[width=1.0\linewidth]{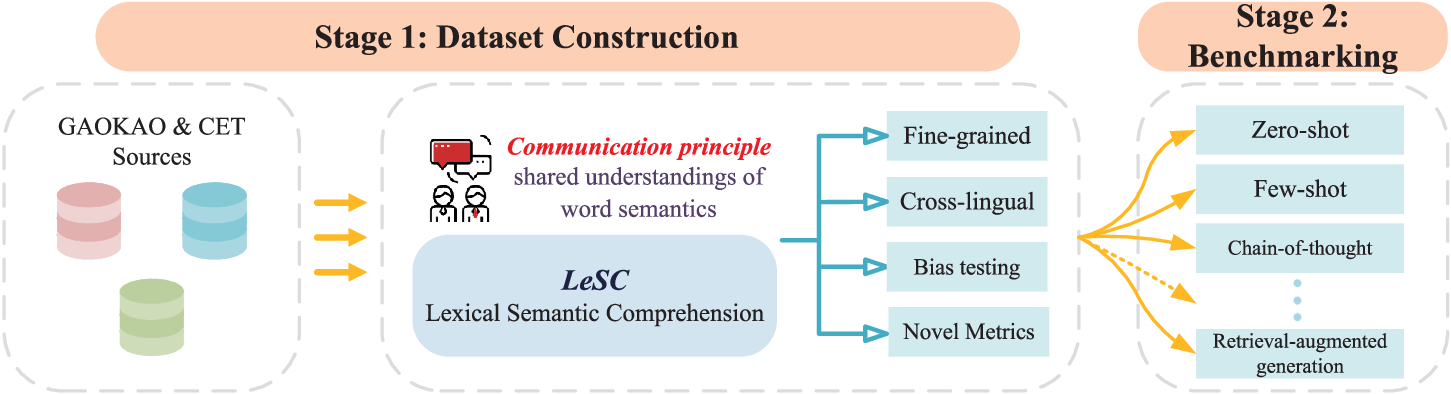}
\caption{The workflow of LeSC. In stage 1, we first construct the LeSC dataset using GAOKAO and CET sources. After that, in stage 2, we employ advanced strategies to obtain benchmarking results for LLMs.}\label{figure2}
\end{figure*}

The paper is organized as follows. Section \textcolor[RGB]{0,128,172}{\ref{sec2}} introduces the constructions of datasets, and models and methods used in this paper. Then experimental results and analysis are presented in section \textcolor[RGB]{0,128,172}{\ref{sec3}}. After that, we show the case study in \textcolor[RGB]{0,128,172}{\ref{sec4}}. Finally, conclusions and future works are discussed in section \textcolor[RGB]{0,128,172}{\ref{sec5}} and \textcolor[RGB]{0,128,172}{\ref{sec6}}.



\section{Materials and methods}
\label{sec2}
In this section, we present the construction process of LeSC benchmark (§\textcolor[RGB]{0,128,172}{\ref{subsec2.1}}) with evaluation details. Selected models and methods (prompting techniques and retrieval-augmented-generation) are also introduced (§\textcolor[RGB]{0,128,172}{\ref{subsec2.2}}) The overall workflow is illustrated in \textcolor[RGB]{0,128,172}{Figure \ref{figure2}}.

\subsection{The LeSC datasets}
\label{subsec2.1}
In this part, we describe the creation process of the LeSC dataset (§\textcolor[RGB]{0,128,172}{\ref{subsubsec2.1.1}}) and the evaluation metrics (§\textcolor[RGB]{0,128,172}{\ref{subsubsec2.1.2}}).

\subsubsection{Dataset creation}
\label{subsubsec2.1.1}

Aiming at measuring LLMs' fine-grained language comprehension capacity in scenarios with low error tolerance, we propose a novel dataset (LeSC) focusing on the lexical level, the first benchmark encompassing both fine-grained and cross-lingual dimensions. Specifically, inspired by datasets originating from standardized testing\textcolor[RGB]{0,128,172}{~\citep{hendryckstest2021,zhang2023evaluating}}, we first collect polysemous words from
the Chinese college entrance examination (GAOKAO) and College English Test (CET-4) online resources, and refine them manually. Then, utilizing online dictionaries\footnote{\url{https://dict.youdao.com/} and \url{https://pinyin.sogou.com/dict/}}, we create a polysemy dictionary linking the above English words with their various Chinese meanings. After that, previous exams, particularly reading comprehension parts, are crawled, and using ChatGPT, irrelevant parts are filtered out to build a 5k sentence dictionary with keys as sentences and values as bilingual word-definition pairs. Finally, after a meticulous human examination and curation process, ambious sentences are removed, resulting in 600 high-quality samples containing a list of multiple semantics for each key word.

Given LLMs' remarkable performance in multiple-choice QA (MCQA) tasks\textcolor[RGB]{0,128,172}{~\citep{hendryckstest2021,pal2022medmcqa,bubeck2023sparks}} like sentiment analysis and text classification (essentially equivalent to MCQA with limited answer options), we have formulated the above-curated samples in the simple multiple-choice format to obtain a fair and objective assessment of language comprehension capabilities. Here, we define the input in LLMs is the combination of a prompt $P$ and a question $x$: $[P, x]$, where $[,]$ denotes the concatenation operation. For $P$, following previous work\textcolor[RGB]{0,128,172}{~\citep{zhu2023promptbench}}, with human design and GPT-4\footnote{\url{https://openai.com/api/}}, we construct prompts for both task-oriented and role-oriented types, each consisting of three samples. For $x$, the order of options remains entirely random, as \textcolor[RGB]{0,128,172}{~\citep{robinson2023leveraging}} highlights LLM's sensitivity while facing a simple change in the order of options (details in \textcolor[RGB]{0,128,172}{Appendix A}). In total, 3600 samples are obtained for LSU evaluation.

Noteworthy, our evaluation encompasses not only the model's understanding of fine-grained lexical semantics but also its capacity for cross-lingual transfer. We have formulated the questions in English while the candidate answers are presented in Chinese (\textcolor[RGB]{0,128,172}{Figure \ref{figure1}} is an example). For comparison, we additionally render the original options in English to facilitate further validation and analysis. More descriptions can be found in \textcolor[RGB]{0,128,172}{Appendix A}.

\subsubsection{Evaluation metrics}
\label{subsubsec2.1.2}
For a question q, we denote its standard answer as $a(q)$, and the answer of LLM $M$ by $M(q)$. To evaluate the comprehension ability of $M$, we consider the following two metrics:

\paragraph{Absolute Accuracy}
This is the most commonly adopted average accuracy in NLP tasks. In the following experiments, unless otherwise specified explicitly, accuracy all refers to this type.

\begin{equation}
    Acc_{abs}(M;Q)=\frac{ {\textstyle \sum_{q\in Q}} \mathbb{I}[M(q)=a(q)]}{|Q|}\label{equ 1}
\end{equation}where $Q$ is the set of all test questions, and $\mathbb{I}$ is the indicator function, which equals 1 if the model answers correctly; otherwise, 0.

\paragraph{Weighted Accuracy}
As illustrated in §\textcolor[RGB]{0,128,172}{\ref{subsubsec2.1.1}}, LLMs exhibit a marked sensitivity to the order of options. That means that for questions correctly answered by the model M, it may be solely due to ground truth labels coincidentally aligning with the option symbols (e.g., 'B') that M tends to favor and select, thus not convincing to indicate M's perfect understanding ability (\textcolor[RGB]{0,128,172}{Appendix A}). Hence, we introduce a novel weighted metric to mitigate impact of model symbol bias:

\begin{equation}
    Acc_{wtd}(M;Q)=\frac{ {\textstyle \sum_{q\in Q}} W(idx(q))\ast \mathbb{I}[M(q)=a(q)]}{ {\textstyle \sum_{w\in W}W} }\label{equ 2}
\end{equation}

\begin{equation}
    W(i) = \frac{1}{acc_{options}({gt(i))} /sum(acc_{options})}\label{equ 3}
\end{equation}\vspace{0.05em}where $W$ denotes the weight matrix with shape $(1,|Q|)$, $idx(q)$ and $gt(i)$ are the index of question $q$ in LeSC and its label, and $acc_{options}$ consists of five elements, with each element representing the absolute performance of M while the ground truth answer for each question in $Q$ is placed at the corresponding element location. Additionally, if a question has fewer than five options ($n < 5$), we maintain its original order of options when computing the value of $acc_{options}$ for positions beyond $n$.

\subsection{Models and methods}
\label{subsec2.2}
In this part, we first describe the models utilized in our research (§\textcolor[RGB]{0,128,172}{\ref{subsubsec2.2.1}}). Then human evaluation baseline alongside random baseline is presented for comparative analysis (§\textcolor[RGB]{0,128,172}{\ref{subsubsec2.2.2}}). Additionally, we delve into advanced strategies, such as prompting techniques (§\textcolor[RGB]{0,128,172}{\ref{subsubsec2.2.3}}) and retrieval-augmented generation (§\textcolor[RGB]{0,128,172}{\ref{subsubsec2.2.4}}), to offer a more comprehensive analysis. Finally, attention visualization is also introduced to give some intuitive explanations (§\textcolor[RGB]{0,128,172}{\ref{subsubsec2.2.5}}).

\subsubsection{Selected models}
\label{subsubsec2.2.1}
To ensure the efficacy and generalizability of our research for both academic researchers and commercial applications, we consider models with varying architectures and sizes, and both open-source and closed-source: GPT-3.5 and GPT-4, Vicuna-v1.5 (7B, 13B, 33B)\textcolor[RGB]{0,128,172}{~\citep{chiang2023vicuna}}, Llama2 (7B, 13B)\textcolor[RGB]{0,128,172}{~\citep{touvron2023llama}}, Qwen (7B, 14B)\textcolor[RGB]{0,128,172}{~\citep{qwen}}, Baichuan2 (7B, 13B)\textcolor[RGB]{0,128,172}{~\citep{baichuan2023baichuan2}}, and ChatGLM3-6B\textcolor[RGB]{0,128,172}{~\citep{du2022glm}}. This allows for a comprehensive quantification of LLMs' language understanding capabilities across various dimensions. We give some brief description below. For more details, please refer to official websites or the corresponding Huggingface repository\footnote{\url{https://huggingface.co/models}}.
\begin{itemize}
    \item \textbf{ChatGPT}: Developed by OpenAI, ChatGPT is a large language model designed to produce human-like text in response to given prompts. Built on the GPT-3 architecture, the GPT-3.5 series has undergone fine-tuning for enhanced interactivity and conversational capabilities. Notably, GPT-4 stands out as the most proficient LLM in performance.

    \item \textbf{Vicuna-v1.5}: The Vicuna model, derived from fine-tuning the LLaMA-2 base model by LMSYS, was developed using around 70K user-shared conversations obtained from ShareGPT.com through public APIs.

    \item \textbf{Llama2}: The Llama2 model, developed by Meta AI's FAIR team, is a widely-used autoregressive language model based on the Transformer architecture.

    \item \textbf{Qwen}: Proposed by Alibaba Cloud, Qwen series are strong base language models, which have been stably pretrained for up to 3 trillion tokens of multilingual data with a wide coverage of domains, languages (with a focus on Chinese and English), etc. They can achieve competitive performance on benchmark datasets.

    \item \textbf{Baichuan2}: Baichuan 2 is the new generation of large-scale open-source language models launched by Baichuan Intelligence. It is trained on a high-quality corpus with 2.6 trillion tokens and has achieved excellent performance in authoritative Chinese and English benchmarks of the same size.

    \item \textbf{ChatGLM3}: ChatGLM3 is the latest open-source model in the ChatGLM series by Zhipu. While retaining many excellent features such as smooth dialogue and low deployment threshold from the previous two generations, ChatGLM3 introduces the following features: more powerful base model, more comprehensive function support.
\end{itemize}

\subsubsection{Human evaluation and random baseline}
\label{subsubsec2.2.2}
\paragraph{Humam evaluation} Following \textcolor[RGB]{0,128,172}{\citep{jang2023can}}, we provide human performance on LSU task to explore the gap between SOTA LLMs and humans. We randomly selected 300 samples from LeSC and then conducted evaluations on 16-year-old humans drawn randomly from international high schools, representing diverse backgrounds, ensuring reliable results. This age is selected based on the widely acknowledged commonsense that sixteen is a milestone marking a significant transition from childhood to adulthood, with the emergence of mature cognitive abilities observed across various cultures\textcolor[RGB]{0,128,172}{~\citep{cowie2019birth,Gordon2022}}. Our study aims to precisely measure the understanding level of LLMs compared to our hypothesis that even 16-year-old humans excel in the simple LSU task. Therefore, if SOTA GPT-4 struggles to surpass human performance, it suggests significant shortcomings in LLMs' language understanding from the perspective of human intelligence.

\paragraph{Random baseline} The number of options in LeSC ranges from 2 to 5, leading to an average of 4.39 across all samples. Here, the random selection level are defined as the reciprocal of this average (22.77$\%$), which we intuitively hold the belief that existing splendid LLMs will undoubtedly outperform.  

\subsubsection{Prompting methods}
\label{subsubsec2.2.3}
Researchers have proposed various prompting methods to inspire inherent energy for improved performance in LLMs. Earlier explorations \textcolor[RGB]{0,128,172}{~\citep{brown2020language,bubeck2023sparks}} have revealed the noteworthy influence of few-shot prompting on enhancing LLM performance. Chain of thought (CoT) prompting\textcolor[RGB]{0,128,172}{~\citep{wei2022chain,kojima2022large}}, a type of few-shot prompting, achieved notable improvements across challenging benchmarks by suggesting a straightforward solution: changing the answers in few-shot examples to step-by-step answers, especially when combined with very large language models like ChatGPT\textcolor[RGB]{0,128,172}{~\citep{open2023introducing}}. Furthermore, \textcolor[RGB]{0,128,172}{~\citep{kojima2022large}}, demonstrated that suitable hints in prompts also result in a respectable performance, even in the absence of any exemplar. However, \textcolor[RGB]{0,128,172}{~\citep{shi2023large}} also pointed out that inappropriate prompts might meanwhile impede the model's comprehension and decision-making processes to some extent. In this paper, we investigate the impact of these cutting-edge prompting techniques on our benchmark LeSC to explore their real effects on the LSU problem.

\subsubsection{Retrieval-augmented generation}
\label{subsubsec2.2.4}
While LLMs have demonstrated impressive capabilities, they face issues such as hallucination and reliance on outdated knowledge\textcolor[RGB]{0,128,172}{~\citep{huang2023survey,kandpal2023large}}. Retrieval-augmented generation has emerged as a promising solution these years\textcolor[RGB]{0,128,172}{~\citep{gao2023retrieval,zhao2024retrieval}}. By integrating pertinent external knowledge and information, this method facilitates accurate and credible generated content and reasoning process, particularly in knowledge-intensive tasks. Earlier work use retrieval models to obtain relevant documents, and directly feed into the generation model with original input\textcolor[RGB]{0,128,172}{~\citep{chen-etal-2017-reading,qu-etal-2021-rocketqa}}. Recent explorations focus on enhancing the performance of retrievers and how to only utilize useful information in retrieved documents\textcolor[RGB]{0,128,172}{~\citep{kandpal2023large,asai2024selfrag,yoran2024making}}. For example, \textcolor[RGB]{0,128,172}{\citep{luo-etal-2023-search}} fine-tunes a language model on the Alpaca instruction-tuning data with top retrieved passages inserted before instructions. GenRead\textcolor[RGB]{0,128,172}{~\citep{yu2023generate}} generates contextual information with the implicit knowledge of LLM, and then produce the final answers with these relevant documents. Self-RAG\textcolor[RGB]{0,128,172}{~\citep{asai2024selfrag}} designs four special tokens and then generates instruction-tuning data labeled by GPT-4, which is then utilized to fine-tune a Llama2. 

In this work, given the unique attributes of our research, retrieving relevant documents from external databases like Wikipedia or conventional search engines may be challenging and inefficient. Furthermore, leveraging ChatGPT's extensive training and learning from over 40TB of internet data, we believe that ChatGPT is adept at serving as an external knowledge source. Thus, given an input $x$ consisting of task instructions and input content, we begin by using ChatGPT to generate relevant documents $d$, followed by extracting and summarizing useful snippets to create the final evidence passages $p$. Ultimately, for LLM $M$, the answer $A$ is produced as follows:
\begin{equation}
    A = M(x,p)\label{equ 4}
\end{equation}

\begin{figure*}[htp!]
\centering
\includegraphics[width=1.0\linewidth]{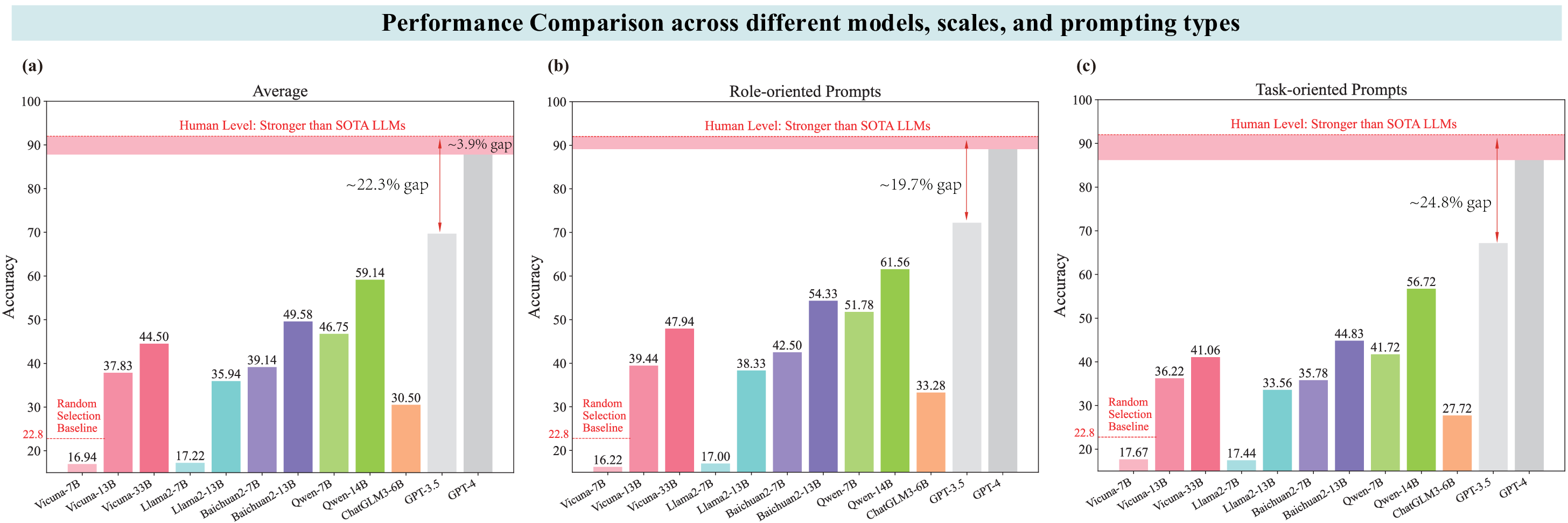}
\caption{Results for the overall performance on LeSC dataset under different settings. Tile 'Average', 'Role-oriented Prompts', and 'Task-oriented Prompts' refer to the accuracy ($\times$ 100) of LLMs on all, role-oriented, task-oriented prompts, respectively. We also plot the performance levels of humans (92$\%$) and random selection (23$\%$) as a reference.}\label{figure3}
\end{figure*}

\subsubsection{Visualization Technique}
\label{subsubsec2.2.5}
Text visualization \textcolor[RGB]{0,128,172}{~\citep{cao2016overview,zhu2023promptbench}} can intuitively and effectively convey information. To evaluate LLM's recognition capabilities on fine-grained semantic information within the sentence (word or phrase), we introduce an attention visualization technique here, i.e. Attention by Gradient, which assigns scores based on gradient norms for each word. 

As an effective method utilized in open-source models, attention by gradient aims to determine the significance of each word in the input through gradient analysis. Specifically, given an input $x$ consisting of $k$ words and $n$ tokens labeled as $y$, we express it as $x=[t_{1}^{1},t_{1}^{2},...,t_{k}^{n}]$, where $t_{i}^{j}$ denotes the $i$-th word and $j$-th token. Thus, tokens corresponding to the same word should be concatenated using a mapping function $w_{i} = f_{map}(t_{i}^{j})$. For the model M, we first compute the token-level gradient as follows:

\begin{equation}
    g_{t_{i}^{j}} = \frac{\partial L(f_{M}(x),y) }{\partial t_{i}^{j}}\label{equ 5}
\end{equation}\vspace{0.05em}where $L$ is the loss function (cross-entropy default in our paper), $f_{M}$ refers to the function of model M.

Upon obtaining the above gradient, word-level gradient can be computed by summing the token-level gradients corresponding to each word as follows:
\begin{equation}
    g_{w_i} = \sum_{j\in 0,1,...,n} g_{t_{i}^{j}}  \quad\text{s.t. }w_{i} = f_{map}(t_{i}^{j})\label{equ 6}
\end{equation}

Finally, we calculate the $l_2$ norm of each word’s gradient, subsequently applying normalization to yield a score, denoted as $s_{w_i}$ for each word:
\begin{equation}
    s_{w_i} = \frac{||g_{w_i}||^{2} - \min g_{w_i}}{\max g_{w_i} - \min g_{w_i}}\label{equ 7}
\end{equation}  

It should be noted that the magnitude of attention scores reflects the model's focus on specific words or nuanced semantics. In other words, the higher attention scores, the correspondingly greater the color depth in visualizations.

\section{Experimental results and analysis}
\label{sec3}
In this section, we provide comprehensive experiments and anticipate answering three challenging yet meaningful questions: §\textcolor[RGB]{0,128,172}{\ref{subsec3.1}} whether the LSU issue exists in SOTA LLMs, §\textcolor[RGB]{0,128,172}{\ref{subsec3.2}} how to mitigate, and §\textcolor[RGB]{0,128,172}{\ref{subsec3.3}} why it persists and proves difficult to resolve. We implement all experiments in this paper with an Nvidia A100 GPU.

\subsection{Are LLMs proficient in lexical semantic understanding?}
\label{subsec3.1}
\begin{figure}[htp!]
\centering
\includegraphics[width=1.0\linewidth]{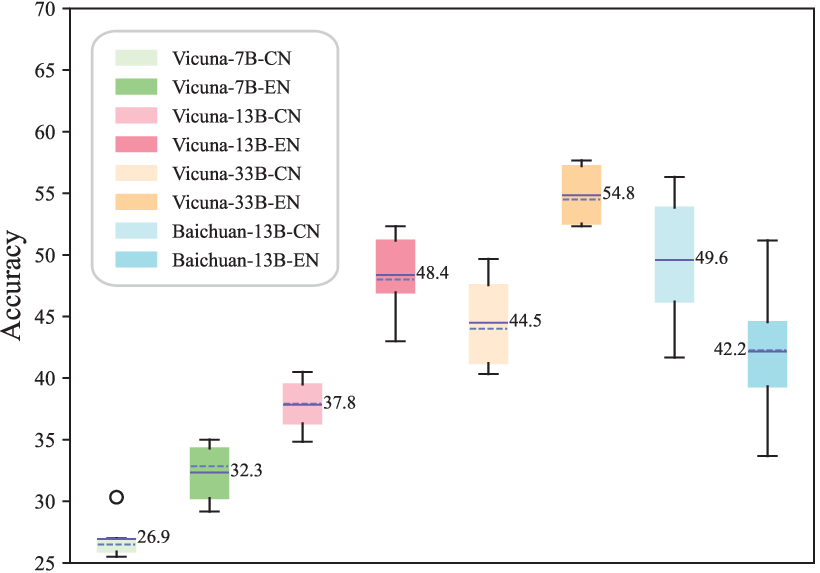}
\caption{Accuracy ($\times$ 100) for different model scales and architectures, and pretraining corpora concerning languages (CN, EN) of options.}\label{figure4}
\end{figure}

We comprehensively compare model performances on LeSC with the main results illustrated in \textcolor[RGB]{0,128,172}{Figure \ref{figure3} and \ref{figure4}}, and detailed results are listed in \textcolor[RGB]{0,128,172}{Appendix B}. Each reported value is the absolute accuracy on the whole dataset under different settings (prompt types, model architectures, sizes, etc.). Our results offer three key insights as follows.

\subsubsection{LLMs including GPT-4 perform poorly on LSU}
\label{subsubsec3.1.1}
As shown in \textcolor[RGB]{0,128,172}{Figure \ref{figure3}}, all considered cutting-edge LLMs, both open-source and closed-source, with diverse scales and architectures, consistently exhibit less-than-satisfactory performances far below expectations on the LSU task. Specifically, as one of the most powerful LLMs in open-source domains\textcolor[RGB]{0,128,172}{~\citep{liu2023agentbench}}, the 30B vicuna model surprisingly achieves less than 50$\%$ accuracy on this task. At the same time, vicuna-7B exhibits a complete lack of understanding, with a performance of 16.94$\%$ that is even worse than random selections. Furthermore, despite slightly better improvements from instruction fine-tuning on llama families and better generalization to new instructions in zero-shot scenarios, vicuna-family models still fall short in this fundamental comprehension task. This suggests that the emergence of this issue could not be attributed to LLMs' difficulty in adhering to instructions\textcolor[RGB]{0,128,172}{~\citep{jang2023can}}, but rather stems from essential challenges in intrinsic fine-grained comprehension.

For larger-scale commercial models, SOTA GPT-4 exhibits an average performance gap of 3.9$\%$ compared with 16-year-old humans while GPT-3.5 leads to 22.3$\%$ performance gap, which contrasts starkly with their near-perfect performance on diverse semantic understanding tasks like reading comprehension\textcolor[RGB]{0,128,172}{~\citep{bubeck2023sparks}}. This thought-provoking discrepancy prompts us to reflect on whether LLMs truly possess fine-grained accurate human-like language comprehension and analytical skills rather than simply relying on co-occurrence contrasts or mimicing language patterns, i.e. stochastic parrots.

\subsubsection{Role-oriented prompts outperform task-oriented}
\label{subsubsec3.1.2}
Despite the overall suboptimal performance, a comparison between the two prompt types (\textcolor[RGB]{0,128,172}{Figure \ref{figure3}}) illustrates a certain excellence in role-oriented prompts over task-oriented ones. This could be attributed to the inherent properties of LLMs in emulating acquired knowledge and task types, enabling effective handling of various problems. Thus, when endowed with role-specific instructions, LLMs may swiftly and accurately navigate the relevant state space, thereby fully leveraging their intrinsic comprehension powers.

\subsubsection{Cross-lingual transfer capacity of LLMs encounters limitations}
\label{subsubsec3.1.3}
Given the original questions and options are set in English and Chinese, respectively, this configuration inadvertently assesses the language capability transfer in LLMs. For in-depth analysis, we employ GPT-4 to convert all options into English, subsequently creating a new dataset named LeSC-EN. As depicted in \textcolor[RGB]{0,128,172}{Figure \ref{figure4}}, partial test results on LeSC-EN are presented, leading us to the following findings. Note that we have transformed the metric value of Vicuna-7B-CN (+10$\%$) for better presentation, which, nevertheless, doesn't influence our conclusions.

Firstly, for models of the same types (Vicuna-CN), we observe a progressive augmentation tendency in the transfer effect of cross-lingual comprehension abilities as the scale ranges from 7B to 33B. However, this trend remains limited to some extent and exhibits a poor correlation with the higher inference costs that come with model scaling. Secondly, disparities in pre-training corpora may account for substantial divergences in cross-lingual transfer among different models. Taking the Vicuna and Baichuan models as an example, Baichuan2-13B, benefiting from a larger share of Chinese pre-training corpora, surpasses Vicuna-13B in the original LeSC task. Nevertheless, upon switching to English (LeSC-EN) for testing, the Vicuna models, endowed with a greater proportion of English corpora, manifest a pronounced performance improvement (about 10$\%$), whereas the Baichuan model undergoes a relative decline in performance.

Our results reveal notable constraints in the cross-lingual transfer proficiency of current LLMs. Consequently, we advocate for an in-depth assessment of models' capabilities and attributes across diverse transfer scenarios, including domains, cultures, and linguistic styles. We consider this imperative for better advancements in LLM research and leave further exploration for future work.

\begin{table*}[htp]
    \centering
    \caption{Absolute and weighted accuracies ($\times$100) with prompting techniques on the LeSC dataset. `Modified', `Ins-hint', `Ins-false', and `Ins-false-hint' are all instructed prompting, which denotes more easy problems after modification, instruction with correct understanding hints, false information, false information plus correct hints like `Feel free to ignore false information in the inputs', respectively. Results with bold and underlining labels are better than others. Here we consider the vicuna-family models with three sizes, 7B, 13B, 33B.}
    \label{table1}
    \begin{tabular*}{\textwidth}{@{\extracolsep{\fill}}llllllll@{\extracolsep{\fill}}}
        \toprule
        \multirow{2.5}{*}{\textbf{Method}}             
        & \multirow{2.5}{*}{\textbf{Explanation}} 
        & \multicolumn{3}{c}{\textbf{Absolute Accuracy}} & \multicolumn{3}{c}{\textbf{Weighted Accuracy}}  \\
        \cmidrule(r){3-5} \cmidrule(r){6-8}
        \quad & \quad & \textbf{7B}  & \textbf{13B} & \textbf{33B} & \textbf{7B}  & \textbf{13B} & \textbf{33B}\\
        \midrule
        Base  & /     & 16.94   & 37.83   & 44.50   & 15.96   & 35.49   & 41.49\\
        Random & \textit{Random selection}  & 22.77   & 22.77   & 22.77   & 22.77   & 22.77   & 22.77\\
        Human  & \textit{Human level} & \textbf{91.90}   & \textbf{91.90}   & \textbf{91.90}   & \textbf{91.90}   & \textbf{91.90}  & \textbf{91.90}\\
        Few-shot    & In-context learning & $32.36_{\textcolor[RGB]{0,176,80}{+15.42}}$   & $45.81_{\textcolor[RGB]{0,176,80}{+7.98}}$   & $49.22_{\textcolor[RGB]{0,176,80}{+4.72}}$   & $32.04_{\textcolor[RGB]{0,176,80}{+16.08}}$   & $45.11_{\textcolor[RGB]{0,176,80}{+9.62}}$   & $49.80_{\textcolor[RGB]{0,176,80}{+8.31}}$\\
        CoT    & \textit{CoT} & $31.67_{\textcolor[RGB]{0,176,80}{+14.73}}$   & $51.03_{\textcolor[RGB]{0,176,80}{+13.20}}$   & $35.14_{\textcolor[RGB]{255,0,0}{-9.36}}$   & $31.30_{\textcolor[RGB]{0,176,80}{+15.33}}$   & $51.87_{\textcolor[RGB]{0,176,80}{+16.38}}$   & $37.35_{\textcolor[RGB]{255,0,0}{-4.14}}$\\
        0-CoT  & \textit{Zero-shot CoT} & $23.83_{\textcolor[RGB]{0,176,80}{+6.89}}$   & $42.00_{\textcolor[RGB]{0,176,80}{+4.17}}$   & $47.81_{\textcolor[RGB]{0,176,80}{+3.31}}$   & $22.44_{\textcolor[RGB]{0,176,80}{+6.48}}$   & $38.84_{\textcolor[RGB]{0,176,80}{+3.35}}$   & $43.81_{\textcolor[RGB]{0,176,80}{+2.32}}$\\
        RAG  & \textit{/} & $\underline{49.78}_{\textcolor[RGB]{0,176,80}{+32.84}}$   & $\underline{53.67}_{\textcolor[RGB]{0,176,80}{+15.84}}$   & $\underline{62.81}_{\textcolor[RGB]{0,176,80}{+18.31}}$   & $\underline{49.00}_{\textcolor[RGB]{0,176,80}{+33.04}}$   & $\underline{52.28}_{\textcolor[RGB]{0,176,80}{+16.79}}$   & $\underline{59.76}_{\textcolor[RGB]{0,176,80}{+18.27}}$\\
        Modified & \textit{Modified problems}   & $29.81_{\textcolor[RGB]{0,176,80}{+12.87}}$   & $52.14_{\textcolor[RGB]{0,176,80}{+14.31}}$   & $59.31_{\textcolor[RGB]{0,176,80}{+14.81}}$   & $27.74_{\textcolor[RGB]{0,176,80}{+11.78}}$   & $49.51_{\textcolor[RGB]{0,176,80}{+14.02}}$   & $56.81_{\textcolor[RGB]{0,176,80}{+15.32}}$\\
        Ins-hint  & \textit{Inst.+hint} & $18.69_{\textcolor[RGB]{0,176,80}{+1.75}}$   & $44.89_{\textcolor[RGB]{0,176,80}{+7.06}}$   & $46.69_{\textcolor[RGB]{0,176,80}{+2.19}}$   & $17.57_{\textcolor[RGB]{0,176,80}{+1.61}}$   & $41.93_{\textcolor[RGB]{0,176,80}{+6.44}}$   & $43.30_{\textcolor[RGB]{0,176,80}{+1.81}}$ \\
        Ins-false      & \textit{Inst.+false info.}           & $6.00_{\textcolor[RGB]{255,0,0}{-10.94}}$   & $10.36_{\textcolor[RGB]{255,0,0}{-27.47}}$   & $1.50_{\textcolor[RGB]{255,0,0}{-43.00}}$   & $5.75_{\textcolor[RGB]{255,0,0}{-10.21}}$   & $8.67_{\textcolor[RGB]{255,0,0}{-26.82}}$   & $1.47_{\textcolor[RGB]{255,0,0}{-40.02}}$\\
        Ins-false-hint & \textit{Inst.+false info.+hint}           & $13.72_{\textcolor[RGB]{255,0,0}{-3.22}}$   & $15.03_{\textcolor[RGB]{255,0,0}{-22.80}}$   & $2.50_{\textcolor[RGB]{255,0,0}{-42.00}}$   & $12.71_{\textcolor[RGB]{255,0,0}{-3.25}}$   & $12.84_{\textcolor[RGB]{255,0,0}{-22.65}}$   & $2.27_{\textcolor[RGB]{255,0,0}{-39.22}}$\\
        \bottomrule
    \end{tabular*}
\end{table*}

\subsection{Can prompting techniques and retrieval-augmented-generation help mitigate?}
\label{subsec3.2}
To investigate the efficacy of advanced strategies in mitigating the issues mentioned in §\textcolor[RGB]{0,128,172}{\ref{subsec3.1}}, we adopt a wide range of prompting approaches described in §\textcolor[RGB]{0,128,172}{\ref{subsubsec2.2.3}}, including zero-shot prompting (Zero-shot), few-shot in-context learning (ICL)\textcolor[RGB]{0,128,172}{~\citep{brown2020language}}, chain-of-thought prompting (CoT)~\citep{wei2022chain}, zero-shot chain-of-thought (0-CoT)\textcolor[RGB]{0,128,172}{~\citep{kojima2022large}} here. Notably, for CoT, to maintain acceptable time cost and prevent over-fitting in prompt engineering, we align with\textcolor[RGB]{0,128,172}{~\citep{shi2023large}} on exemple creation; that means, we only utilize one simple exemple. For 0-CoT, we follow \textcolor[RGB]{0,128,172}{\citep{kojima2022large}} and directly present the candidate problem followed by ``Answer: Let’s think step by step:". Additionally, we incorporate manually designed prompts to explicitly guide LLMs towards adherence to human intuitions, commonly termed instructed prompting. For retrieval-augmented-generation (RAG), the overall process is illustrated in §\textcolor[RGB]{0,128,172}{\ref{subsubsec2.2.4}}. Three notable findings can be discerned as follows.

\begin{figure}[!ht]
\centering
\includegraphics[width=1.0\linewidth]{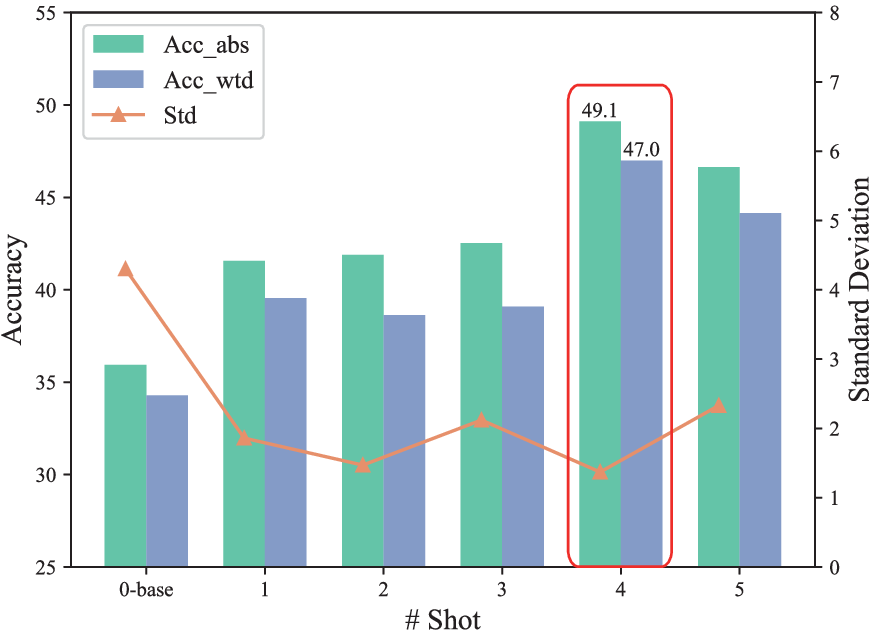}
\caption{Results on LeSC corresponding to k (shot) in few-shot prompting. '$Acc_{a}$ ', '$Acc_{wtd}$', 'Std' refer to absolute and weighted accuracy, and standard deviation, respectively.}\label{figure5}
\end{figure}

\subsubsection{ICL enhances performance within a certain range}
\label{subsubsec3.2.1}
As illustrated in \textcolor[RGB]{0,128,172}{Figure \ref{figure5}} on Llama2-13B, several key conclusions could be derived. Firstly, in terms of both stability and two specified metrics, incorporating appropriate examples (few-shot) consistently surpasses zero-shot scenarios.  While prior work has shown that prompting with exemplars improves model robustness\textcolor[RGB]{0,128,172}{~\citep{shi2023large}}, our work indicates that in-context learning also results in notable comprehension improvement. This can be attributed to its efficacy in enhancing the model's ability to capture contextual information adeptly, thereby facilitating a more comprehensive understanding of the input context and user intent. Secondly, with the increase in the number of examples (up to 4), the model performance exhibits a gradual upward trend, which can be explained by the extensive pre-training of LLMs on massive data, enabling the internalization of diverse language structures and knowledge. The introduction of additional examples serves to better activate LLMs' inherent prior knowledge, coupled with contextual information, fostering a deeper understanding of the context and thereby effectively accomplishing NLU tasks. However, an excess of examples may introduce noise, potentially impairing the model's generalizability and decision-making processes in novel scenarios. More results are shown in \textcolor[RGB]{0,128,172}{Appendix B}.

\subsubsection{RAG enhances model semantic understanding, especially for small models}
\label{subsubsec3.2.2}
As shown in \textcolor[RGB]{0,128,172}{Table~\ref{table1}}, introducing external knowledge contributes to enhancing LLMs' comprehension capacities for nuanced semantics, especially for small models. For example, for vicuna-7B, the performance increased by 194$\%$, indicating that RAG can enhance semantic understanding. However, its effectiveness is still limited, and the impact of model sizes has been removed to some extent. In other words, the final model performance largely depends on the effectiveness of retrieved content.

\subsubsection{Scaling challenges: diminishing benefits of advanced prompting in large language models}
\label{subsubsec3.2.3}

\begin{figure}[htp!]
\centering
\includegraphics[width=1.0\linewidth]{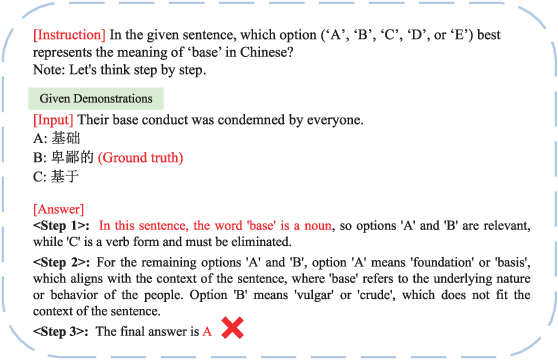}
\caption{An example from LeSC dataset that showcases the CoT result of Vicuna-33B. Option  'A', 'B', 'C' refer to 'the part on which it rests or is supported', 'contemptibly low in position', 'to use as a foundation', respectively.}\label{figure6}
\end{figure}

\textcolor[RGB]{0,128,172}{Table~\ref{table1}} illustrates that advanced prompting techniques, such as CoT and instructed prompting, effectively enhance overall performance on LeSC. However, on very large LLMs, these strategies might be ineffective or even counterproductive. Our observations reveal a consistent inverse scaling law for methods like CoT and few-shot in-context learning, diminishing their efficacy in improving LLMs' performance. Notably, on the Vicuna-33B model, CoT has unexpectedly resulted in detrimental effects. We conjecture that as the model scale increases, the biases in the built-in pre-training corpus escalate, leading to overconfidence in its existing prior knowledge when processing tasks. Simultaneously, larger models, calibrated with human intent\textcolor[RGB]{0,128,172}{~\citep{ouyang2022training}}, may exhibit more heightened adherence to human instructions. In such cases, additional cues like CoT may conflict with LLM's intrinsic knowledge, causing uncertainty, potential hallucinations, and ultimately impacting decision-making accuracy. 

Taking \textcolor[RGB]{0,128,172}{Figure \ref{figure6}} as an example, the question is to comprehend and infer the meaning of the word 'base' within the sentence "Their base conduct was condemned by everyone." Employing the Cot method for stepwise analysis, the model is initially required to perform part-of-speech (POS) tagging to eliminate inappropriate options. While LLMs typically excel in POS tagging tasks\textcolor[RGB]{0,128,172}{~\citep{chang2023survey,bubeck2023sparks}}, the Vicuna-33B model, as outlined above, anomalously identifies "base" as a noun, deviating from the anticipated outcome.

\begin{table*}[htp]
    \centering
    \caption{Response analysis based on attention visualization. The green and red colors in 'pred' denote right and wrong answers, respectively. Color intensity denotes attention weights (heavier color means larger weights). False information ('Ins-false') results in incorrect responses. Even with correct ignoring hints ('Ins-false-hint'), LLMs directly disregard them, remaining influenced by misleading information."}
    \label{table3}
    \begin{tabular*}{\textwidth}{llp{15cm}lp{15cm}}
        \toprule
        \textbf{Type} & \hspace*{11pt}\textbf{Pred} & \hspace*{11pt}\textbf{Inputs}  \\
        \midrule
        Base              &  \hspace*{11pt}\textcolor[RGB]{0,176,80}{B}    & \hspace*{11pt}\colorbox[RGB]{252,168,139}{Input:\vphantom{fg}}\hspace*{0pt}\colorbox[RGB]{252,173,145}{I\vphantom{fg}}\hspace*{0pt}\colorbox[RGB]{252,180,153}{don't\vphantom{fg}}\hspace*{0pt}\colorbox[RGB]{252,182,156}{follow\vphantom{fg}}\hspace*{0pt}\colorbox[RGB]{252,190,165}{your\vphantom{fg}}\hspace*{0pt}\colorbox[RGB]{246,89,63}{line\vphantom{fg}}\hspace*{0pt}\colorbox[RGB]{252,195,172}{of\vphantom{fg}}\hspace*{0pt}\colorbox[RGB]{239,59,44}{reasoning.\vphantom{fg}}\hspace*{0pt}\colorbox[RGB]{251,129,97}{A:\vphantom{fg}}\hspace*{0pt}\colorbox[RGB]{244,78,56}{line\vphantom{fg}}\hspace*{0pt}\colorbox[RGB]{191,21,26}{segment\vphantom{fg}}\hspace*{0pt}\colorbox[RGB]{244,80,57}{B:\vphantom{fg}}\hspace*{0pt}\colorbox[RGB]{103,0,12}{method\vphantom{fg}}\hspace*{0pt}\colorbox[RGB]{253,209,191}{Answer:\vphantom{fg}}\hspace*{0pt}\\
        \specialrule{0em}{2pt}{2pt}
        Ins-false         &  \hspace*{11pt}\textcolor{red}{A}     & \hspace*{11pt}\colorbox[RGB]{250,0,0}{False Information\vphantom{fg}}\hspace*{0pt}\newline\hspace*{11pt}\colorbox[RGB]{252,162,132}{Input:\vphantom{fg}}\hspace*{0pt}\colorbox[RGB]{252,198,175}{I\vphantom{fg}}\hspace*{0pt}\colorbox[RGB]{252,195,172}{don't\vphantom{fg}}\hspace*{0pt}\colorbox[RGB]{252,192,168}{follow\vphantom{fg}}\hspace*{0pt}\colorbox[RGB]{253,209,191}{your\vphantom{fg}}\hspace*{0pt}\colorbox[RGB]{248,94,66}{line\vphantom{fg}}\hspace*{0pt}\colorbox[RGB]{253,214,197}{of\vphantom{fg}}\colorbox[RGB]{248,97,68}{reasoning.\vphantom{fg}}\hspace*{0pt}\colorbox[RGB]{239,59,44}{A:\vphantom{fg}}\hspace*{0pt}\colorbox[RGB]{228,48,39}{line\vphantom{fg}}\hspace*{0pt}\colorbox[RGB]{103,0,12}{segment\vphantom{fg}}\hspace*{0pt}\colorbox[RGB]{221,41,36}{B:\vphantom{fg}}\hspace*{0pt}\colorbox[RGB]{245,86,61}{method\vphantom{fg}}\hspace*{0pt}\colorbox[RGB]{253,212,194}{Answer:\vphantom{fg}}\hspace*{0pt}\\
        \specialrule{0em}{2pt}{2pt}
        Ins-false-hint    &  \hspace*{11pt}\textcolor{red}{A}     & \hspace*{11pt}\colorbox[RGB]{254,240,233}{The\vphantom{fg}}\hspace*{0pt}\colorbox[RGB]{254,235,226}{following\vphantom{fg}}\hspace*{0pt}\colorbox[RGB]{253,219,203}{note\vphantom{fg}}\hspace*{0pt}\colorbox[RGB]{254,239,231}{has\vphantom{fg}}\hspace*{0pt}\colorbox[RGB]{254,242,236}{some\vphantom{fg}}\hspace*{0pt}\colorbox[RGB]{253,223,209}{false\vphantom{fg}}\hspace*{0pt}\colorbox[RGB]{253,223,209}{information,\vphantom{fg}}\hspace*{0pt}\colorbox[RGB]{254,243,237}{and\vphantom{fg}}\hspace*{0pt}\colorbox[RGB]{255,245,240}{just\vphantom{fg}}\hspace*{0pt}\colorbox[RGB]{255,245,240}{feel\vphantom{fg}}\hspace*{0pt}\colorbox[RGB]{254,243,238}{free\vphantom{fg}}\hspace*{0pt}\colorbox[RGB]{254,242,236}{to\vphantom{fg}}\hspace*{0pt}\colorbox[RGB]{254,241,234}{ignore\vphantom{fg}}\hspace*{0pt}\colorbox[RGB]{253,215,199}{them.\vphantom{fg}}\hspace*{0pt}\newline\hspace*{11pt}\colorbox[RGB]{250,0,0}{False Information\vphantom{fg}}\hspace*{0pt}\newline\hspace*{11pt}\colorbox[RGB]{251,141,109}{Input:\vphantom{fg}}\hspace*{0pt}\colorbox[RGB]{252,191,166}{I\vphantom{fg}}\hspace*{0pt}\colorbox[RGB]{252,189,163}{don't\vphantom{fg}}\hspace*{0pt}\colorbox[RGB]{252,180,153}{follow\vphantom{fg}}\hspace*{0pt}\colorbox[RGB]{252,198,175}{your\vphantom{fg}}\hspace*{0pt}\colorbox[RGB]{245,86,61}{line\vphantom{fg}}\hspace*{0pt}\colorbox[RGB]{252,201,180}{of\vphantom{fg}}\hspace*{0pt}\colorbox[RGB]{248,96,67}{reasoning.\vphantom{fg}}\hspace*{0pt}\colorbox[RGB]{196,22,27}{A:\vphantom{fg}}\hspace*{0pt}\colorbox[RGB]{219,39,35}{line\vphantom{fg}}\hspace*{0pt}\colorbox[RGB]{103,0,12}{segment\vphantom{fg}}\hspace*{0pt}\colorbox[RGB]{155,12,19}{B:\vphantom{fg}}\hspace*{0pt}\colorbox[RGB]{247,90,64}{method\vphantom{fg}}\hspace*{0pt}\colorbox[RGB]{252,180,153}{Answer
        :\vphantom{fg}}\hspace*{0pt}\\
        \bottomrule
    \end{tabular*}
\end{table*}

\subsubsection{LLMs significantly focus more on misleading information over corrective instructions}
\label{subsubsec3.2.4}
To further investigate the underlying reasons for the above issues, we employ instructed prompting methods here\textcolor[RGB]{0,128,172}{~\citep{shi2023large}}. By providing both correct and incorrect information, we aim to discern whether LLMs are just stochastic parrots and genuinely lack nuanced semantic understanding or conform to previous research conclusions\textcolor[RGB]{0,128,172}{~\citep{jang2023can}} that `LLMs merely struggle with following instructions, not its comprehension ability'. Specifically, three scenarios are configured: `instruction with correct hint', `instruction with false information', and `instruction with false information and also `correct ignoring hint'. The outcomes reveal a noteworthy impact of these instructions on LLMs, reaching up to 96$\%$ effectiveness. In other words, within the defined scope here, models demonstrate proficiency in following instructions and subsequently executing tasks. Thus, we attribute the subpar performances in §\textcolor[RGB]{0,128,172}{\ref{subsec3.1}} primarily to LLMs' inadequate semantic comprehension.

Additionally, we observe an intriguing finding that, in lexical comprehension tasks, LLMs significantly prioritize misleading information over valuable instructions. More specifically, with the `ins-hint' setup, the benefits of correct instructions are relatively limited. Nevertheless, while introducing erroneous information (`ins-false'), LLMs experience a pronounced decline in performance, exemplified by Vicuna-33B's 96$\%$ drop rate. Even provided with explicit instructions to ignore incorrect information, their dependency on incorrect instructions remains challenging to mitigate. This finding highlights the need for future research to focus more on the impact of inevitable false information and explore practical alleviating strategies. \textcolor[RGB]{0,128,172}{Table~\ref{table3}} is an example using attention visualization illustrated in §\textcolor[RGB]{0,128,172}{\ref{subsubsec2.2.5}}, where LLMs directly disregard the correct instruction `The following note has some false information, and just feel free to ignore them'.

\begin{table}
    \centering
    \caption{Comparison between zero-shot performance and translation process, involving initial translation acquisition for a candidate problem, followed by GPT-4 evaluation for translation accuracy.}
    \label{table2}
    \begin{tabular*}{\linewidth}{@{\extracolsep{\fill}}llll@{\extracolsep{\fill}}}
        \toprule
        \textbf{Model} & \textbf{Size} & \textbf{Acc-Base} & \textbf{Acc-Translation}  \\
        \midrule
        \multirow{3}{*}{Vicuna-v1.5}  &  7B    & 16.94   & $\textbf{48.50}_{\textcolor[RGB]{0,176,80}{+31.56}}$\\
        \quad                         &  13B   & 37.83   & $\textbf{56.50}_{\textcolor[RGB]{0,176,80}{+18.67}}$\\
        \quad                         &  33B   & 44.50   & $\textbf{46.33}_{\textcolor[RGB]{0,176,80}{+1.83}}$\\
        \multirow{2}{*}{Llama-2}      &  7B    & 17.22   & $\textbf{32.67}_{\textcolor[RGB]{0,176,80}{+15.45}}$\\
        \quad                         &  13B   & 35.94   & $\textbf{45.50}_{\textcolor[RGB]{0,176,80}{+9.56}}$\\
        \bottomrule
    \end{tabular*}
\end{table}

\subsection{Conjectures on why this problem exists}
\label{subsec3.3}
In this section, we perform an in-depth analysis to unravel the intricacies of this issue, addressing the question of why this problem is hard to solve for LLMs.

\subsubsection{Inherent limitation of stochastic parrots}
\label{subsubsec3.4.1}
We further supplement an experiment to delve into its underlying mechanisms, since word, phrase, or sentence-level semantic understanding tasks are inherently akin to translation tasks, which LLMs excel at\textcolor[RGB]{0,128,172}{~\citep{chang2023survey}}. Particularly, we guide LLMs to translate the provided questions, subsequently employing GPT-4 to assess the results, examining their effectiveness in conveying nuanced semantic information. That means, for a candidate question, if GPT-4 determines that a translation accurately captures uncommon meanings of common words, we consider it a successful sample.

\begin{figure}[ht!]
\centering
\includegraphics[width=0.95\linewidth]{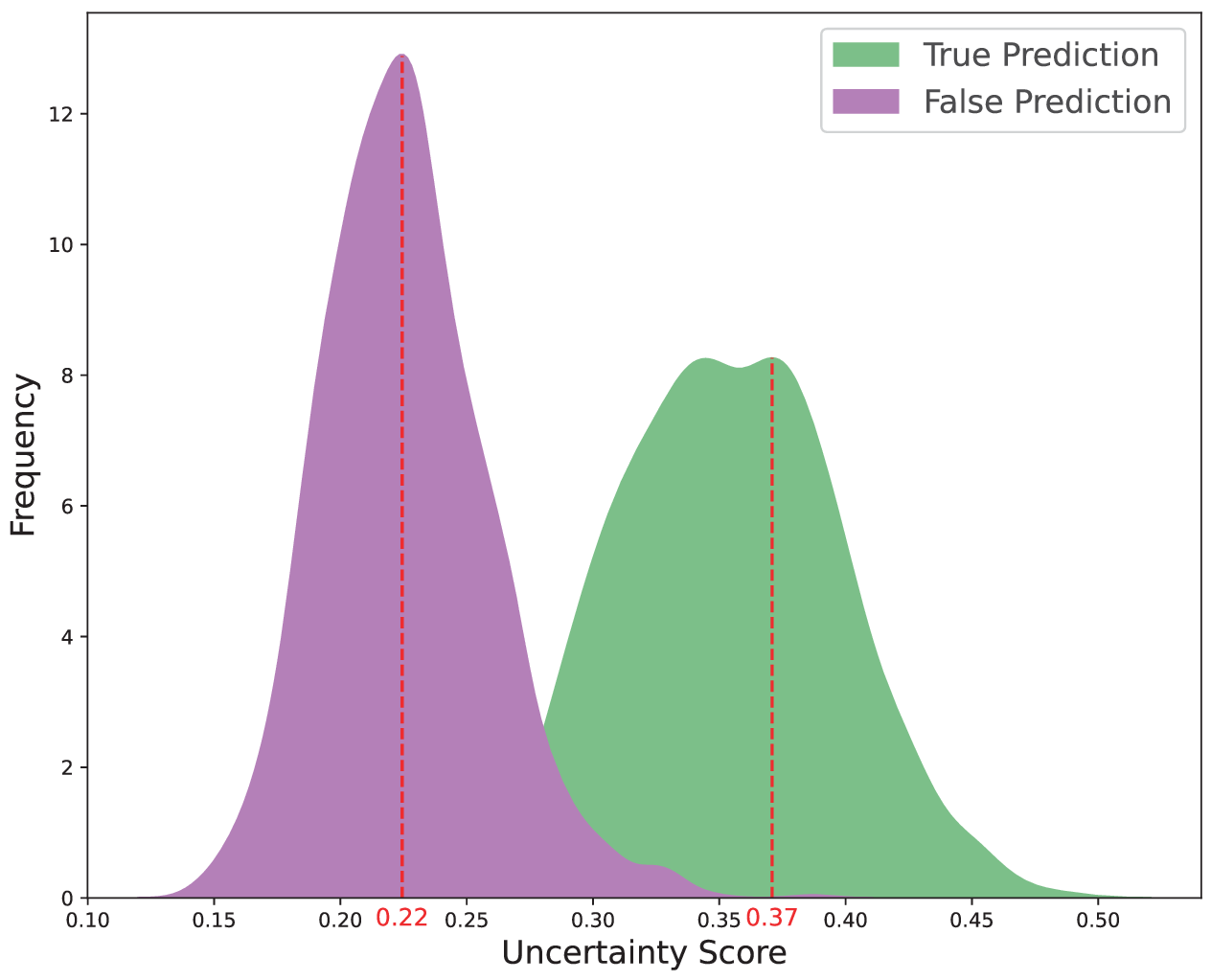}
\caption{Comparison between correctly answered and incorrectly answered questions on uncertainty. The results show that LLMs tend to give false responses while exhibiting low uncertainty and heightened confidence.}\label{figure7}
\end{figure}

As shown in \textcolor[RGB]{0,128,172}{Table \ref{table2}}, while introducing translation steps, we observe a consistent improving trend on models of varying scales (7B, 13B, 33B) and types (instructed-tuning Vicuna-v1.5 and original Llama-2). Given this phenomenon, we argue that, in direct question-answering scenarios, the model's comprehension and cross-lingual transfer capabilities are often constrained by substantial pre-training corpus biases and unfamiliarity with question formats. However, introducing translation steps fundamentally exposes the model to a more familiar and proficient task paradigm, thereby capitalizing on its advantages in co-occurrence contrast and imitation. This produces efficient extraction of relevant information from pre-training knowledge and semantic contextual details from provided questions, leading to more precise answers. Nevertheless, it must be noted that the translation gains appear to diminish with the increasing model scales because of the current constraints on the LLM's sophisticated comprehension. This phenomenon also emphasizes the limited effectiveness of current approaches, highlighting the urgency for further research into more potent solutions. 

\begin{table*}[htp]
    \centering
    \caption{Output example for different models on a LeSC problem and corresponding zero-shot and CoT solutions, which may potentially contain false or hallucination contents. The left answers correspond to zero-shot, while the solutions and answers on the right correspond to CoT. And the black (e.g. C \faCheckCircle) and red colors (e.g. \textcolor{red}{A \faTimesCircle}) denote right and wrong answers, respectively. Note that options 'A: 空气’, ’B: 外观’, ’C: 公开发表’, 'D: 晒干' refer to 'A: the invisible gaseous substance surrounding the earth, a mixture mainly of oxygen and nitrogen', 'B: the appearance or impression given by a person or thing', 'C: to make (a grievance or problem) public', 'D: to dry (something) by exposing it to air'.}
    \label{table5}
    \begin{tabularx}{\textwidth}{lp{1cm}XX}
        \toprule
        \multirow{4}{*}{\textbf{Model}} 
        & \multicolumn{2}{l}{\textbf{Question:}}\\
        \quad & \multicolumn{2}{l}{Prompt: In the given sentence, which option best represents the meaning of 'air' in Chinese?}\\
        \quad & \multicolumn{2}{l}{Input: The weekly meeting enables employees to air their complaints.}\\
        \quad & \multicolumn{2}{l}{A: 空气\quad B: 外观\quad \textcolor[RGB]{82,175,199}{C: 公开发表 (Ground Truth)}\quad D: 晒干}\\
        \midrule
        ChatGLM3-6B     & \textcolor{red}{A \faTimesCircle}   & \textbf{\textless Step 1\textgreater:} In the given sentence, the word 'air' is likely used as a verb meaning to express or share something. So, options 'A', 'B', 'C', and 'D' do not fit this context. \textbf{\textless Step 2\textgreater:} Option 'E' can be eliminated because 'air' is not used as a noun or a verb related to the meeting in this sentence. Therefore, the best option for the Chinese meaning of 'air' in this context is 'A'. \textbf{\textless Step 3\textgreater:} \textcolor{red}{The final answer is A \faTimesCircle}\\
        Llama2-7B       & \textcolor{red}{A \faTimesCircle}   & \textbf{\textless Step 1\textgreater:} In the given sentence, the word 'air' is a verb, so option 'A' (noun) is non-conforming and must be removed, leaving only 'B', 'C', and 'D'. \textbf{\textless Step 2\textgreater:} Then for remaining options ['B', 'C', 'D'], option B refers to the appearance or look of something, which does not align with the context of airing complaints. Option C means to publicly express or reveal something, which is close but not the best fit for the context of airing complaints. Option D means to dry or expose something to the air, which is also not the best fit for the context of airing complaints. Therefore, the best option for the Chinese meaning of 'air' in the sentence is 'B'. \textbf{\textless Step 3\textgreater:} \textcolor{red}{The final answer is B \faTimesCircle}\\
        Llama2-13B      & \textcolor{red}{A \faTimesCircle}   & \textbf{\textless Step 1\textgreater:} In the given sentence, the word 'air' is a noun, so we need to find the option that best represents the meaning of 'air' as a noun. \textbf{\textless Step 2\textgreater:} Options A and D both relate to the physical sensation of air, but option D specifically refers to the sensation of hot air, which is not the meaning of 'air' in the sentence. Option C is a noun, but it does not fit the context of the sentence. \textbf{\textless Step 3\textgreater:} Therefore, the best option for the Chinese meaning of 'air' based on how it is used in this sentence is option B, which refers to the appearance or look of something. \textbf{\textless Step 4\textgreater:} \textcolor{red}{The final answer is B \faTimesCircle}\\
        Vicuna-13B      & \textcolor{red}{A \faTimesCircle}   & \textbf{\textless Step 1\textgreater:} In the given sentence, the word 'air' is a verb, so option 'A' (noun) is non-conforming and must be removed, leaving only 'B', 'C', and 'D'. \textbf{\textless Step 2\textgreater:} Then for remaining options ['B', 'C', 'D'], option B refers to appearance or exterior, which doesn't fit the context of the sentence. Option C means 'to make something public or known', which is not the meaning of 'air' in this sentence. Option D means 'to expose to the air', which is not the meaning of 'air' in this sentence. Therefore, none of the options B, C, or D are suitable for the Chinese meaning of 'air' based on how it is used in this sentence. \textbf{\textless Step 3\textgreater:} \textcolor{red}{Since none of the options are suitable, the answer is 'None of the above' \faTimesCircle}\\
        Baichuan-13B    & \textcolor{red}{D \faTimesCircle}  & \textbf{\textless Step 1\textgreater:} In the provided phrase, "air" means express oneself freely without reservation or fear about what others might say as if you are speaking directly into an open space like freshly cleaned room where dust particles have been cleared out. Option D ("晒") also has similar connotation but usually implies sharing personal experiences/stories publicly. Option C doesn't fit well because it suggests publishing something formally while we need more freedom when using “air” here. \textbf{\textless Step 2\textgreater:} Among these choices, there seems no direct match between English words and corresponding terms in Mandarin except for "air". So let’s check whether any other term can replace them accurately. \textbf{\textless Step 3\textgreater:} \textcolor{red}{Finally, considered above factors, I believe that the correct response should include "air" \faTimesCircle}\\
        Qwen-14B        & C \faCheckCircle   & \textbf{\textless Step 1\textgreater:} In the given sentence, the word 'air' is a verb, so option 'A' (noun) is non-conforming and must be removed, leaving only 'C' and 'D'. \textbf{\textless Step 2\textgreater:} Then for remaining options ['C', 'D'], option C translates to 'publicly express', which aligns with the context of airing complaints. And option D refers to drying, not fitting the meaning of airing complaints in this sentence. Therefore, C is the best choice for the Chinese meaning of 'air' based on how it's used in sentence. \textbf{\textless Step 3\textgreater:} The final answer is C \faCheckCircle\\
        \bottomrule
    \end{tabularx}
\end{table*}

\subsubsection{Overconfidence to some degree}
\label{subsubsec3.4.2}
As discussed in\textcolor[RGB]{0,128,172}{~\citep{si2022prompting,ying2023intuitive}}, LLMs accumulate substantial knowledge within their parameters, and this knowledge capacity scales with model size. Here, we conjecture that LLMs, having undergone pre-training on vast data with billions of parameters, internalize extensive knowledge, thereby fostering heightened self-confidence and occasionally manifesting biased tendencies. In such a hypothesis, confronted with relatively novel tasks that necessitate fine-grained comprehension, even applying advanced techniques such as CoT to structure the problem into familiar subtasks like part-of-speech tagging, these models encounter challenges. Instead, they persistently furnish responses based on their perceived correctness. We believe it will be especially more difficult for large LLMs, since they are powerful language modeling representers, meaning that it would be harder to make them revert original thoughts by focusing on real-scenario problems. Our conjecture can be partially proven by \textcolor[RGB]{0,128,172}{Figure \ref{figure7}}. We randomly collect 1200 samples each of correctly answered and incorrectly answered questions. Subsequently, we extract their outputs from the model's final layer and calculated the uncertainty score as the negative likelihood value. Our results reveal that LLMs frequently make erroneous predictions when exhibiting low uncertainty and heightened confidence. Conversely, LLMs provide correct responses when they demonstrate a more cautious tendency.

\subsubsection{Coarse-Grained Competence: Impaired Detail-Oriented Task Performance}
\label{subsubsec3.4.3}
LLMs exhibit remarkable proficiency in tasks mainly involving macro-level comprehension, such as translation and summarization. However, given their inclination towards imitating and co-reference resolution characteristics, in the process of massive data-driven pre-training, these models may merely map the acquired knowledge to a relatively coarse and sparse latent space, overlooking the profound semantics of fundamental discrete units like words. In other words, although LLMs are regarded as understanding the world well, they exhibit language patterns and logic mimicking characteristics on these fine-grained tasks, which we refer to as `stochastic parrots'. Hence, we propose that future research should delve into analyzing LLMs' understanding of basic logical symbols and other fine-grained elements, aiming to facilitate the construction of more nuanced and enriched world-mapping supermodels.

\section{Case Study}
\label{sec4}
In addition to the example shown in \textcolor[RGB]{0,128,172}{Figure 1}, we include more example problems and predictions by different models and prompting methods (\textcolor[RGB]{0,128,172}{Tables \ref{table5}}). We can observe that even with the implementation of advanced prompting techniques such as CoT, the generation of accurate responses remains challenging due to issues such as hallucination and the accumulation of errors. These challenges may stem from an insufficient capacity for fine-grained semantic comprehension.

\section{Conclusions}
\label{sec5}
Increasing studies highlight LLMs' exceptional performance across complex NLU tasks. To enhance future progress in LLMs, researchers are increasingly exploring their fundamental nature: are they merely stochastic parrots based on probabilistic statistics, or do they truly possess human-like nuanced semantic understanding? This question remains a mystery, lacking fine-grained scientific assessments. To bridge this gap, this paper proposes a brand-new benchmark, LeSC, aiming at exploring the genuine linguistic-cognitive skills of LLMs. Comprehensive empirical results indicate that existing LLMs face significant challenges in accurately capturing nuanced lexical semantic information, even GPT-4 and GPT-3.5 exhibit noticeable disparities from 16-year-old human performance, with gaps of 3.9$\%$ and 22.3$\%$, respectively. Furthermore, we examine various advanced prompting techniques and retrieval-augmented-generation, and reveal their limited alleviating effects. Finally, some explanations and case studies are also discussed. The findings of our work can aid in the development of more intelligent LLMs, and also extend to other AI areas like realistic image generation.

\section{Future directions}
\label{sec6}
In addition to offering novel insights into LLMs' comprehension capabilities, we also emphasize three promising future directions. Firstly, given our findings, we encourage researchers to prioritize addressing this fundamental limitation while devoting efforts towards advancing LLM development. Secondly, incorporating additional models and prompting techniques is recommended to enhance the comprehensiveness of evaluation results, thereby contributing to a more profound understanding of LLMs. Last but not least, beyond the noted deficiency in language cross-lingual transfer, it's worthwhile to further investigate LLMs' transferability in domains such as cross-cultural and cross-linguistic styles, which may pave the way for powerful artificial general intelligence.

\section{Data availability}
\label{sec8}
The datasets analyzed during the current study are available at \href{https://github.com/jinyangwu/LeSC}{https://github.com/jinyangwu/LeSC.}

\section{Acknowledgements}
\label{sec9}
This research was supported by the National Key Research $\&$ Development Plan of China (No. 2023YFC3305903), and the National Natural Science Foundation of China (NSFC) (No. 62322120, No. 62306316).

\bibliographystyle{elsarticle-harv} 
\bibliography{reference}

\end{CJK}
\end{document}